\documentclass[a4paper,10pt]{article}
\usepackage[T1]{fontenc}
\usepackage[latin1]{inputenc}
\usepackage[centerlast,bf]{caption} 
\usepackage{times}
\usepackage{multicol}
\usepackage{amsmath}
\usepackage{amssymb}
\usepackage{graphicx}
\graphicspath{{./figures/}}
\DeclareGraphicsExtensions{.pdf,.png}
\usepackage{epic}
\usepackage{eepic}
\usepackage{epsfig}
\usepackage{xspace}
\usepackage{color}
\usepackage{booktabs}
\usepackage{subfig}
\usepackage{float}
\usepackage[table]{xcolor}
\definecolor{lightgray}{gray}{0.9}

\newenvironment*{mytitle}{\begin{LARGE}\bf}{\end{LARGE}\\[1.5ex]}%
\newenvironment*{myabstract}{\begin{Large}\bf}{\end{Large}\\[2.5ex]}%
\makeatletter
{}

\makeatother

\begin{document}

\begin{mytitle} High Quality Large-Scale 3-D Urban Mapping with Multi-Master TomoSAR \end{mytitle}

Yilei Shi, \,  Technische Universit{\"a}t M{\"u}nchen (TUM-LMF), yilei.shi@tum.de, Germany\\
Richard Bamler, \,  German Aerospace Center (DLR) \& Technische Universit{\"a}t M{\"u}nchen (TUM-LMF), Germany\\
Yuanyuan Wang, \,  Technische Universit{\"a}t M{\"u}nchen (TUM-SIPEO), Germany\\
Xiao Xiang Zhu, \,  German Aerospace Center (DLR) \& Technische Universit{\"a}t M{\"u}nchen (TUM-SIPEO), Germany\\


\begin{myabstract} Abstract \end{myabstract}
\textcolor{blue}{This is the pre-acceptance version, to read the final version please go to IEEE XPlore.} Multi-baseline interferometric synthetic aperture radar (InSAR) techniques are effective approaches for retrieving the 3-D information of urban areas. In order to obtain a plausible reconstruction, it is necessary to use large-stack interferograms. Hence, these methods are commonly not appropriate for large-scale 3-D urban mapping using TanDEM-X data where only a few acquisitions are available in average for each city. This work proposes a new SAR tomographic processing framework to work with those extremely small stacks, which integrates the non-local filtering into SAR tomography inversion. The applicability of the algorithm is demonstrated using a TanDEM-X multi-baseline stack with 5 bistatic interferograms over the whole city of Munich, Germany. Systematic comparison of our result with airborne LiDAR data shows that the relative height accuracy of two third buildings is within two meters, which outperforms the TanDEM-X raw DEM. The promising performance of the proposed algorithm paved the first step towards high quality large-scale 3-D urban mapping.

\vspace{4ex}	

\section{Introduction}
TanDEM-X satellite is a German civil and commercial high-resolution synthetic aperture radar (SAR) satellite which has almost identical configuration as its 'sister' TerraSAR-X satellite. Together with TerraSAR-X, they are aiming to provide a global high-resolution digital elevation model (DEM) \cite{bib:krieger2007tandem}. Since its launch in 2010, TanDEM-X has been continuously providing high quality bistatic interferograms that are nearly free from deformation, atmosphere and temporal decorrelation.

Tomographic synthetic aperture radar (TomoSAR) is a cutting-edge SAR interferemetric technique that is capable of reconstructing the 3-D information of scatterers and retrieving the elevation profile. Among the many multi-baseline InSAR techniques, TomoSAR is the only one that strictly reconstructs the full reflectivity along the third dimension elevation. SAR tomography and its differential form (D-TomoSAR) have been extensively developed in last two decades \cite{bib:reigber2000first}\cite{bib:gini2002layover}\cite{bib:fornaro2005three}\cite{bib:zhu2010very}\cite{bib:ge2018spaceborne}. They are excellent approaches for reconstructing the urban area and monitoring the deformation, especially when using high resolution data like TerraSAR-X  \cite{bib:zhu2012demonstration} \cite{bib:zhu2013tomo} or COSMO-Skymed \cite{bib:fornaro2014multilook}. Compare to the classic multi-baseline InSAR algorithms, compressive sensing (CS) based methods  \cite{bib:zhu2010tomographic} \cite{bib:budillon2011three} can obtain extraordinary accuracy for TomoSAR reconstruction and show the super-resolution (SR) power, which is very important for urban areas, since layover is dominant.

Although TanDEM-X bistatic data has many advantages, there is only a limited number of acquisitions available for most areas. For a reliable reconstruction, SAR tomography usually requires fairly large interferometric stacks ($> \textrm{20}$ images), because the variance of the estimates is asymptotically related to the product of SNR and the number of acquisitions. Therefore, it is not appropriate for the micro-stacks, which have limited number of interferograms \cite{bib:zhu2012super}.

In this work, we extend the concept of non-local compressive sensing TomoSAR in \cite{bib:shi2018non} and propose a new framework of spaceborne multi-baseline SAR tomography with TanDEM-X bistatic micro-stacks, i.e. 3 to 5 interferograms, which is applicable for real global 3-D urban mapping.

\section{Non-Local TomoSAR for Multi-Master InSAR}
\begin{figure}
  \centering
  \includegraphics[width=0.45\textwidth]{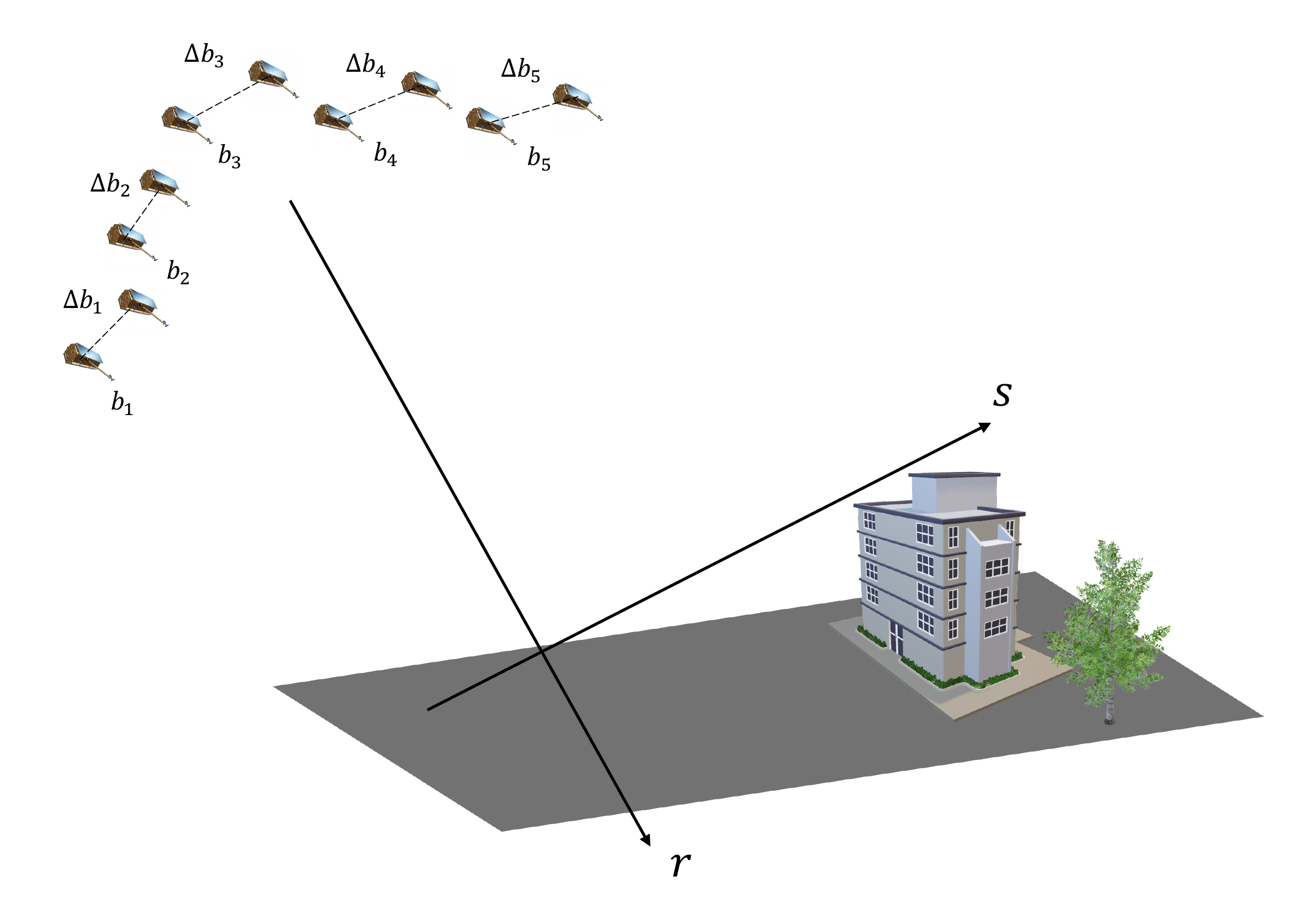}
  \caption{Illustration of multi-master multi-baseline SAR imaging.}
  \label{fig:illu_multi_master}
\end{figure}
In this section, we introduce the non-local TomoSAR framework for multi-master multi-baseline InSAR configuration. Fig. \ref{fig:illu_multi_master} illustrates multi-master multi-baseline SAR imaging. The framework consists of several steps: (1) non-local filtering; (2) spectral estimation; (3) model selection; (4) robust height estimation.

For a fixed azimuth-range position, $\gamma(s)$ represents the reflectivity profile along elevation $s$. The measurement $g_n$ in the $n^{th}$ SAR image is then a sample of $\Gamma(k)$ -- the Fourier transform of $\gamma(s)$, where the elevation wavenumber $k$ is a scaled version of the sensor's position $b_n$ projected on the cross-range-azimuth axis $b||s$ :
\begin{equation}
  g_n = \Gamma(k_n) = \int \gamma(s) \exp(-jk_ns)ds
\end{equation}
with
\begin{equation}
  k_n = -\dfrac{4 \pi b_n}{\lambda r}
\end{equation}
Note that $b_n$ are no baselines, but the positions of the sensor w.r.t. some origin. In case of monostatic multi-temporal data stacks, a single master $g_0$ is chosen with $b_0 = 0$ and interferograms to all other acquisitions are formed : $g_ng_0^*$.

Here we are dealing with stacks of \textit{bistatic} acquisitions, i.e. with the multi-master case. From each of these acquisitions we get a master $g_{n,m} = \Gamma(k_n)$ taken at $b_{master} = b_n$ and a slave $g_{n,s} = \Gamma(k_n+\Delta k_n)$ image taken at $b_{slave} = b_n + \Delta b_n$, where $\Delta b_n$ is the bistatic baseline (which takes the \textit{effective} positions of the transmit-receive phase center into account). The consequence is, that we cannot simply throw this data stack into a standard, i.e. single-master, TomoSAR inversion algorithm and thus confuse $\Delta b_n$ and $b_n$. If there is only a single scatterer in $\gamma(s)$, this misinterpretation would do no harm, because the Fourier transform of a single point has a constant magnitude and a linear phase. In order to determine the slope of the phase ramp we can take any two samples and divide their phase difference by the difference in wavenumbers (= baseline). This is no longer true for two or more scatterers. The example of two symmetric and equally strong scatterers makes this clear:
\begin{gather}
  \gamma(s) = \delta(s+s_0) + \delta(s-s_0) \nonumber \\
  \updownarrow \\
  \Gamma(k) = 2\cos(s_0k)= 2\cos(2\pi \dfrac{2s_0}{\lambda r}b) \nonumber
\end{gather}
Hence, interferograms with the same baseline $\Delta b$ are different depending on where the two sensors were located along $b$. If by chance one of the sensors is at a zero of $\Gamma(k)$, e.g. at $b = \lambda r / 8 s_0$, the interferogram would be zero. Obviously, every bistatic acquisition provides three pieces of information: the two magnitudes $|\Gamma(k_n)|$ and $|\Gamma(k_n+\Delta k_n)|$ as well as the phase difference $\angle \Gamma(k_n+\Delta k_n)\Gamma^*(k_n)$ which have to be accounted for by the inversion algorithm.

This is true for pixel-wise tomographic inversion or for point scatterers. The situation becomes different, though, once we talk about averages of pixels, i.e. estimates of expectation values. Let us assume Gaussian distributed scattering with a backscatter coefficient along elevation of
\begin{equation}
  \sigma_0 = E \left\{ |\gamma(s)|^2 \right\}
\end{equation}
Assuming further that $\gamma(s)$ is white, its power spectral density is stationary and is the autocorrelation function of $\Gamma(k)$, i.e. the Fourier transform of $\sigma_0(s)$ as a function of the baseline wavenumber $\Delta k$ :
\begin{equation}
  E \left\{ \Gamma(k_n+\Delta k_n) \Gamma^*(k)\right\} = \int \sigma_0(s)\exp(-j\Delta k_n s)ds
\end{equation}
Instead of sampling the Fourier spectrum we sample its autocorrelation function by the bistatic data stack. Since this relationship is \textit{independent} of $k \propto b$ because of stationarity, it makes no difference, where the two acquisitions have been taken, only their baseline $\Delta b_n$ counts. In other words we can use standard TomoSAR inversion algorithms in this case.

In this paper we use nonlocal filtering to improve SNR for micro-stacks. These filters perform ensemble averages with number of looks in the order of tens to hundreds. Hence, we tend to the assumption that we work with reasonably good estimates of $E \left\{ \Gamma(k_n+\Delta k_n) \Gamma^*(k)\right\}$ and use the \textit{bistatic interferograms} for TomoSAR reconstruction. By introducing a noise $\boldsymbol{\varepsilon}$, the matrix notation of TomoSAR model can be formulated as:
\begin{equation}
\mathbf{g}=\mathbf{R}\mathbf{X}+\boldsymbol{\varepsilon}
\label{equ:tomosar_basic}
\end{equation}
where $\mathbf{g} = [g_1, g_2, ..., g_n]^{\mathrm{T}}$ is vector notation of the complex-valued measurement with dimension $N \times 1$, and $\mathbf{X} \sim \sigma_0(s_l) = E\{|\gamma(s_l)|^2\}$ is the expectation value of reflectivity profile along elevation uniformly sampled at $s_l (l=1,2,...,L)$. $\mathbf{R}$ is a sensing matrix with the dimension $N \times L$, where $R_{nl} = \exp(-j\Delta k_ns_l)$.

\subsection{Non-Local Procedure}
Since we have only limited number of acquisitions for large-scale area, the SNR need to be dramatically increased in order to obtain the required accuracy. As shown in \cite{bib:shi2018non}, non-local procedure is an efficient way to increase the SNR of interferograms without notable resolution distortion. The idea of patch-wise non-local means considers all the pixels $s$ in the search window, when the patch with the central pixel $s$ is similar to the patch with central pixel $c$, the value of $s$ is selected for calculating the value of pixel $c$. The value of pixel $c$ is estimated by using a weighted maximum likelihood estimator (WMLE).
\begin{equation}
\hat{\boldsymbol{\Theta}}_c = \mathrm{argmax} \sum_s \mathbf{w}(i_s, j_s) \log p(\mathbf{g}_s|\boldsymbol{\Theta})
\end{equation}
where weights $\mathbf{w}(i_s, j_s)$ can be calculated by using patch-wise similarity mesurement \cite{bib:shi2018non}. Assuming that we have two expressions $\mathbf{g} = (I_1, I_2, \phi)$ and $\boldsymbol{\Theta} = (\psi, \mu, \sigma^2)$, where $\mathbf{g}$ denotes the complex-valued measurement. $I_1$ and $I_2$ are the instensity of two SAR images. $\phi$ is the interferometric phase.

\begin{figure*}
  \centering
  \subfloat[]{\includegraphics[width=0.4\textwidth]{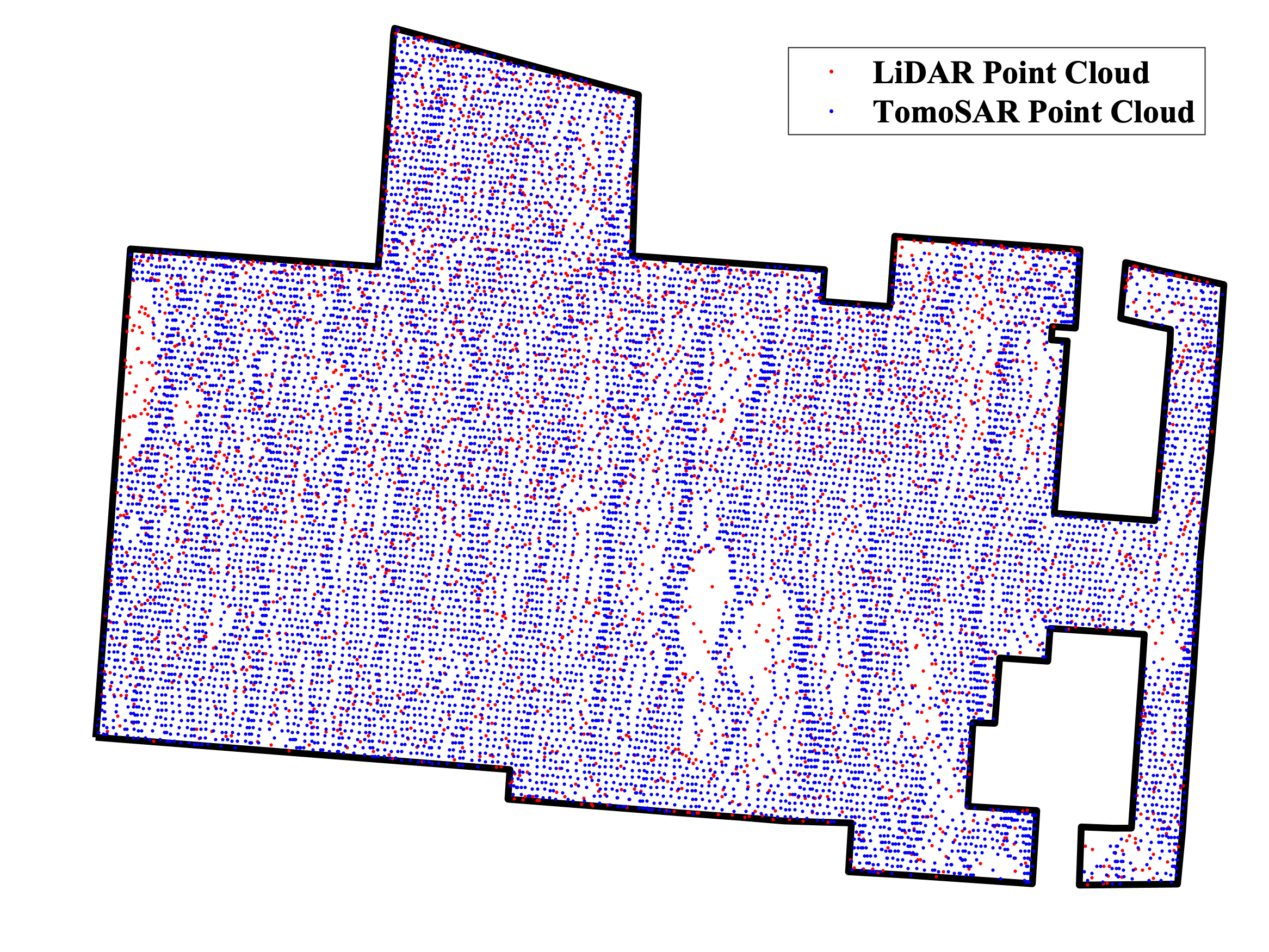}}
  \hfil
  \subfloat[]{\includegraphics[width=0.4\textwidth]{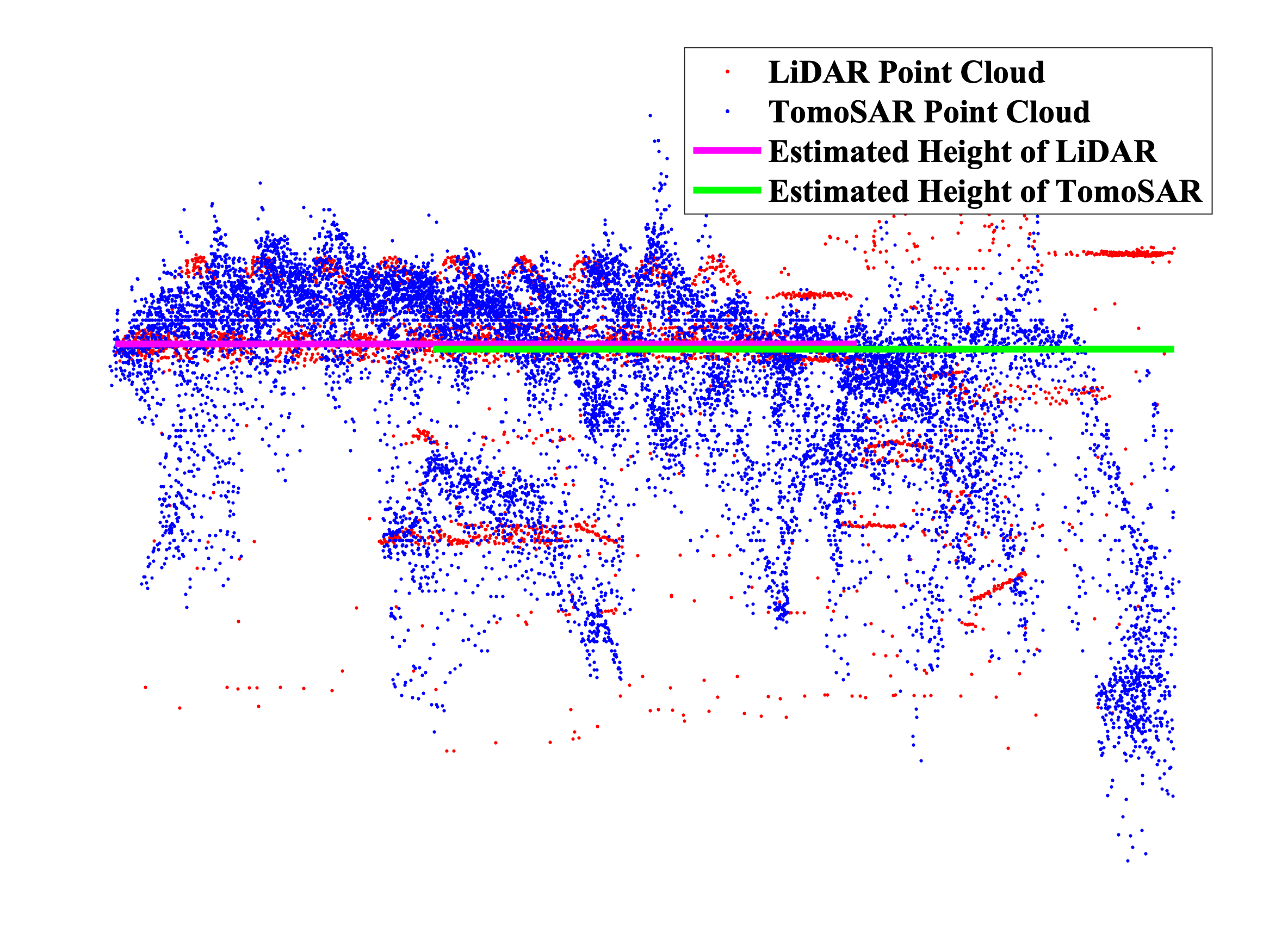}}
  \caption{Example of Robust Height Estimation of LiDAR and TomoSAR point clouds. (a) Top view of two point clouds, i.e., LiDAR (red) and TomoSAR (blue). Note that since LiDAR is too dense, only $1\%$ is visuallized. (b) robust height estimation of two point clouds, LiDAR point cloud (red dots), TomoSAR point cloud (blue dots), estimated building height of LiDAR data (magenta solid line), estimated building height of TomoSAR data (green solid line).}
  \label{fig:optical_comp9}
\end{figure*}

\subsection{Spectral Estimation}
After the non-local procedure, spectral estimation is applied. The most relevant spectral estimation algorithms, including singular value decomposition (SVD) \cite{bib:fornaro2005three} \cite{bib:zhu2010very}, compressive sensing (CS) are introduced in the following.
\begin{itemize}
\item SVD:
\begin{equation}
  \hat{\mathbf{X}} = \left( \mathbf{R}^{\textrm{H}} \mathbf{C}_{\varepsilon\varepsilon}^{-1} \mathbf{R} +  \mathbf{C}_{XX}^{-1} \right)^{-1} \mathbf{R}^{\textrm{H}} \mathbf{C}_{\varepsilon\varepsilon}^{-1} \mathcal{N}(\mathbf{g})
\end{equation}

\item CS:
\begin{equation}
\hat{\mathbf{X}} = \arg \min_{\mathbf{X}} \{ \Vert \mathbf{R}\mathbf{X} - \mathcal{N}(\mathbf{g}) \Vert^2_2 + \lambda \Vert \mathbf{X} \Vert_1 \}
\label{equ:opt_nll1lsp}
\end{equation}
\end{itemize}
We follow the procedure proposed in \cite{bib:wang2014efficient}. The elevation profile is first estimated by an efficient low-order spectral estimator in order to discriminate the number of scatterers in one resolution cell. Then, CS-based approach is adopted for the pixel which has multiple scatterers. This method decreases the amount of pixels that need the $L_1$ minimization, which leads to reduce the computational cost. Furthermore, the rest of pixels can be efficiently solved by randomized blockwise proximal gradient method \cite{bib:shi2018fast}.

\subsection{Robust Height Estimation}
To tackle the possible remaining outliers in the height estimates, the final height will be fused from the result of multiple neighbouring pixels as a post-processing. But instead of simple averaging, the height will be adjusted robustly using an \textit{M-estimator}. Instead of minimizing the sum of squared residuals in averaging, M-estimator minimizes the sum of a customized function $\rho\left(\centerdot\right)$ of the residuals:
\begin{equation}
\tilde{s}=\mathop{\arg}\underset{\textit{s}}{\mathop{\min}}\sum\limits_{i}{\rho\left(\hat{s}_i- s\right)},
\label{eq:M_estimator}
\end{equation}
where $\hat{s}_i$ is the elevation estimates of the $i$th neighbouring pixel. It is shown that the close-formed solution of Eq. (\ref{eq:M_estimator}) is simply a weighted averaging of the heights of the neighbouring pixels. The weighting function can be expressed as follows, if the derivative of $\rho\left(x\right)$ exists.
\begin{equation}
w\left(x\right)=\frac{\partial\rho\left(x\right)}{x\partial x}
\end{equation}

\section{Practical Demonstration}

\subsection{Data Description}
We make use of a stack of five co-registered TanDEM-X bistatic interferograms to evaluate the proposed algorithm. The dataset is over Munich, Germany, whose slant range resolution is 1.2 m and the azimuth resolution is 3.3 m. The images were acquired from July 2016 to April 2017.

\subsection{Quantitative Validation}
In this section, we have quantitatively compared the TomoSAR point clouds with precise LiDAR reference. The LiDAR dataset of Munich is provided by Bavarian State Office for Survey and Geoinformation with ten centimeter accuracy. As different data sources have different coordinates and quality, we apply the following steps on the data. (1) Geocoding of TomoSAR point cloud; (2) Co-registration of different point clouds; (3) Object-based raster data generation; (4) Robust height estimation. Here we show an example of these pre-processing steps for the structure "munich central station" in Fig. \ref{fig:optical_comp9}.

\begin{figure}[!ht]
  \centering
  \includegraphics[width=0.4\textwidth]{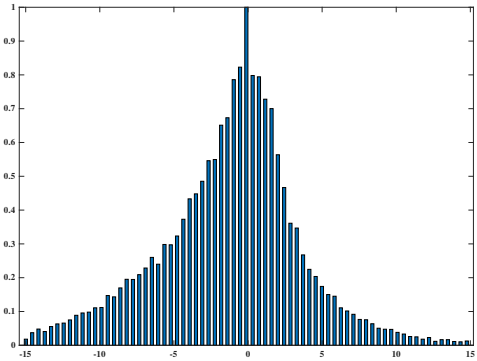}
  \caption{Histogram of height differences of structures.}
  \label{fig:comp_munich_hist}
\end{figure}

In order to have an assessment of the overall accuracy in a city scale, we compared all the 36,499 buildings in the area with the LiDAR point cloud. 38.7\% buildings are within 1 m accuracy. 62.8\% are within 2 m accuracy. However, the two datasets (TanDEM-X CoSSC and LiDAR) were acquired at different time. It is almost certain that changes happened during the period. Therefore, in order to obtain a more realistic assessment, we truncated the distribution of height difference at $\pm \textrm{15 m}$. The truncated histogram can be seen in Fig. \ref{fig:comp_munich_hist}. 34,054 buildings remains after the truncation. Their overall standard deviation is 1.96 m.

\subsection{Fusion with Building Footprint}
\begin{figure}[!ht]
  \centering
  \includegraphics[width=0.45\textwidth]{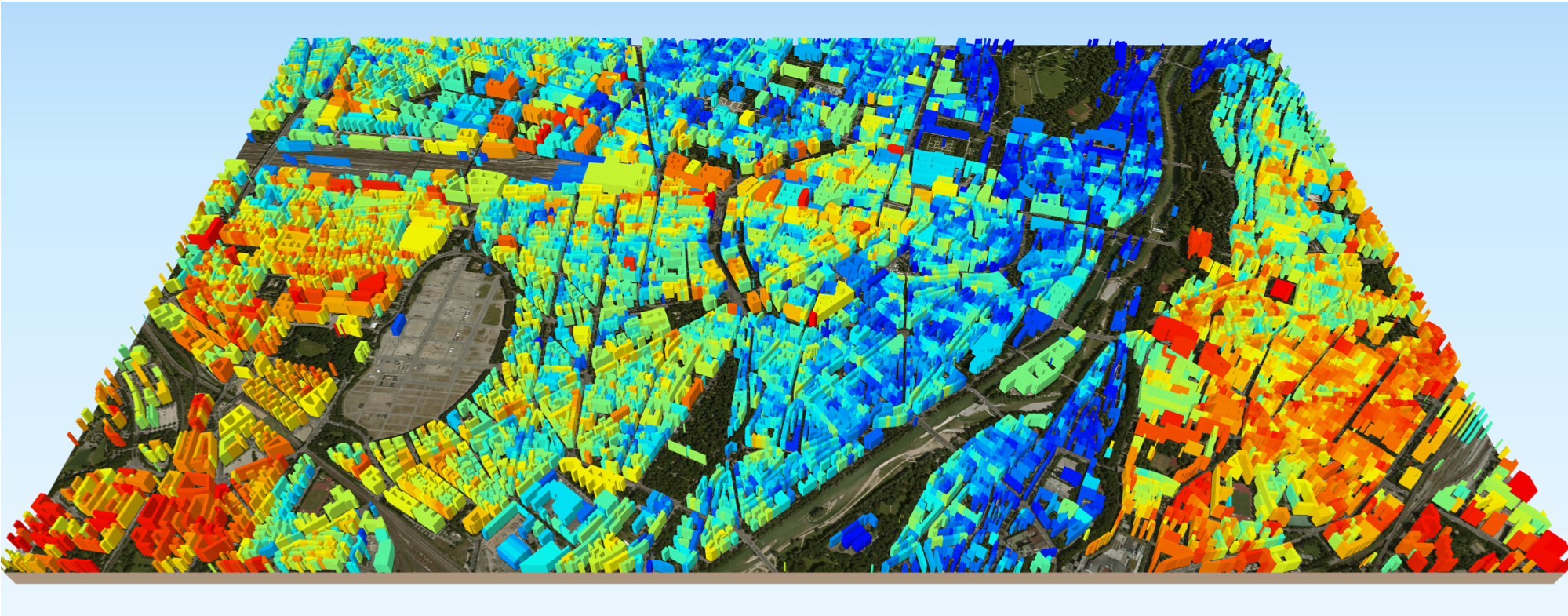}
  \caption{Visualization of fused 3-D urban model of Munich.}
  \label{fig:3d_model_webgl}
\end{figure}

Finally, LOD1 polyhedral models are generated. The 3-D urban models are reconstructed by extruding OpenStreetMap (OSM) with the building height estimated by the proposed multi-master non-local TomoSAR approach. Fig. \ref{fig:3d_model_webgl} shows the fused 3-D urban model of Munich. Color indicates the height of the buildings and 3-D models are overlayed on the Google Map images.

\section{Conclusion}
A new SAR tomographic inversion framework tailored for very limited number of measurements is proposed in this paper. Experiments using TanDEM-X bistatic data shows the relative height accuracy of 2 m can be achieved in large scale. Thus it demonstrates the proposed framework being a promising solution for high quality large-scale 3-D urban mapping.



\begin{thebibliography}{9}\leftskip1mm\advance\labelsep\leftskip
  \bibitem{bib:krieger2007tandem}
  G. Krieger, A. Moreira, H. Fiedler, I. Hajnsek, M. Werner, M. Younis, and M. Zink, ``TanDEM-X: A satellite formation for high-resolution SAR interferometry,'' \emph{IEEE Trans. Geosci. Remote Sens.}, vol. 45, no. 1, pp. 3317-3341, Oct. 2007.

  \bibitem{bib:reigber2000first}
  A. Reigber, and A. Moreira, ``First demonstration of airborne SAR tomography using multibaseline L-band data,'' \emph{IEEE Trans. Geosci. Remote Sensing}, vol. 38, no. 5, pp. 2142-2152, Sep. 2000.

  \bibitem{bib:gini2002layover}
  F. Gini, F. Lombardini, and M. Montanari, ``Layover solution in multibaseline SAR interferometry,'' \emph{IEEE Trans. Aerosp. Electron. Syst.}, vol. 38, no. 4, pp. 1344-1356, 2002.

  \bibitem{bib:fornaro2005three}
  G. Fornaro, F. Lombardini, and F. Serafino, ``Three-dimensional multipass SAR focusing: experiments with long-term spaceborne data,'' \emph{IEEE Trans. Geosci. Remote Sensing}, vol. 43, no. 4, pp. 702-714, Apr. 2005.

  \bibitem{bib:zhu2010very}
  X. X. Zhu and R. Bamler, ``Very high resolution spaceborne SAR tomography in urban environment,'' \emph{IEEE Trans. Geosci. Remote Sens.}, vol. 48, no. 12, pp. 4296-4308, Dec. 2010.

  \bibitem{bib:ge2018spaceborne}
  N. Ge, F. Rodriguez Gonzalez, Y. Wang, Y. Shi, and X. X. Zhu,  ``Spaceborne Staring Spotlight SAR Tomography--A First Demonstration With TerraSAR-X,'' \emph{IEEE J. Sel. Top. Appl. Earth Obs. Remote Sens.}, vol. 11, no. 10, pp. 702-714, Oct. 2018.

  \bibitem{bib:zhu2012demonstration}
  X. X. Zhu and R. Bamler, ``Demonstration of super-resolution for tomographic SAR imaging in urban environment,'' \emph{IEEE Trans. Geosci. Remote Sens.}, vol. 50, no. 8, pp. 3150-3157, Aug. 2012.

  \bibitem{bib:zhu2013tomo}
  X. X. Zhu and Y. Wang and S. Gernhardt and R. Bamler, ``Tomo-GENESIS: DLR's tomographic {SAR} processing system,'' \emph{Proc. Joint Urban Remote Sensing Event, 2013} pp. 159-162.

  \bibitem{bib:fornaro2014multilook}
  G. Fornaro, A. Pauciullo, D. Reale, and S. Verde. ``Multilook SAR Tomography for 3-D Reconstruction and Monitoring of Single Structures Applied to COSMO-SKYMED Data,'' \emph{IEEE J. Sel. Top. Appl. Earth Obs. Remote Sens.}, vol. 7, no. 7, pp. 2776-2785, Jul. 2014.

  \bibitem{bib:zhu2010tomographic}
  X. X. Zhu and R. Bamler, ``Tomographic SAR inversion by $L_{1}$ norm regularization -- The compressive sensing approach,'' \emph{IEEE Trans. Geosci. Remote Sens.}, vol. 48, no. 10, pp. 3839-3846, Oct. 2010.

  \bibitem{bib:budillon2011three}
  A. Budillon, A. Evangelista, and G. Schirinzi, ``Three-dimensional SAR focusing from multipass signals using compressive sampling,'' \emph{IEEE Trans. Geosci. Remote Sens.}, vol. 49, no. 1, pp. 488-499, Jan. 2011.

  \bibitem{bib:zhu2012super}
  X. X. Zhu and R. Bamler, ``Super-resolution power and robustness of compressive sensing for spectral estimation with application to spaceborne tomographic SAR,'' \emph{IEEE Trans. Geosci. Remote Sens.}, vol. 50, no. 1, pp. 247-258, Jan. 2012.

  \bibitem{bib:shi2018non}
  Y. Shi, X. X. Zhu and R. Bamler, ``Non-Local Compressive Sensing Based SAR Tomography,'' \emph{IEEE Trans. Geosci. Remote Sens.}, vol. 57, no. 5, pp.  3015-3024, May 2019.

  \bibitem{bib:wang2014efficient}
  Y. Wang, X. X. Zhu, and R. Bamler, ``An Efficient tomographic inversion approach for urban mapping using meter resolution SAR image stacks,'' \emph{IEEE Geosci. Remote Sens. Lett.}, vol. 11, no. 7, pp. 1250-1254, Jul. 2014.

  \bibitem{bib:shi2018fast}
  Y. Shi, X.X. Zhu, W. Yin, and R. Bamler, ``A fast and accurate basis pursuit denoising algorithm with application to super-resolving tomographic SAR,'' \emph{IEEE Trans. Geosci. Remote Sens.}, vol. 56, no. 10, pp. 6148-6158, Oct. 2018.
\end{thebibliography}
\end{document}